# Anti-Money Laundering Systems Using Deep Learning

MASHKHAL ABDALWAHID SIDIQ[1], YIMAMU KIRUBEL WONDAFEREW[1]

[1]School of Software Engineering, Nankai University, China

*Abstract*: In this paper, we focused on using deep learning methods for detecting money laundering in financial transaction networks, in order to demonstrate that it can be used as a complement or instead of the more commonly used rule based systems and conventional Anti-Money Laundering (AML) systems. The paper explores the pivotal role played by Anti-Money Laundering (AML) activities in the global financial industry. It underscores the drawbacks of conventional AML systems, which exhibit high rates of false positives and lack the sophistication to uncover intricate money laundering schemes. To tackle these challenges, the paper proposes an advanced AML system that capitalizes on link analysis using Deep Learning techniques. At the heart of this system lies the utilization of centrality algorithms like Degree Centrality, Closeness Centrality, Betweenness Centrality, and PageRank. These algorithms enhance the system's capability to identify suspicious activities by examining the influence and interconnections within networks of financial transactions The significance of Anti-Money Laundering (AML) efforts within the global financial sector is discussed in this paper. It highlights the limitations of traditional AML systems. The results showed the practicality and superiority of the new implementation of the GCN model with it being a preferable method for connectively structured data, meaning that a transaction or account is analyzed in the context of its financial environment. In addition, the paper delves into the prospects of Anti-Money Laundering (AML) efforts, proposing the integration of emerging technologies such as Deep Learning and Centrality Algorithms. This integration holds promise for enhancing the effectiveness of AML systems by refining their capabilities.

## Introduction

The document is organized into four main parts. The first is the introductory part, which is the overview of the Money laundering and Anti- money laundering system. The second part of the document covers the existing system and the problems or challenges of current system technology. The third part, Dataset and data preprocessing Techniques, Methodologies, and implementation details. Additionally, we will discuss the potential impact and implications of this research for the field of AML, emphasizing the critical need for more advanced and sophisticated approaches in combating money laundering in today's complex financial landscape.

## Background

The background of this research paper lies in the critical importance of Anti-Money Laundering (AML) efforts within the global financial industry. Money laundering poses a significant threat to the integrity and stability of financial systems worldwide. To combat this illicit activity, financial institutions, and regulatory bodies have implemented AML systems. However, these conventional systems have several limitations that hinder their effectiveness. One major drawback of conventional AML systems is the high rate of false positives. These systems often generate a large number of alerts for potentially suspicious activities, most of which turn out to be legitimate transactions. These false positives overwhelm AML professionals and divert their attention away from genuine money laundering activities. Furthermore, conventional systems lack the sophistication to uncover intricate money laundering schemes that involve complex networks and connections. To address these challenges and improve the effectiveness of AML efforts, this research paper proposes an advanced AML system that utilizes link analysis using a Deep algorithm. Link analysis focuses on exploring the relationships and connections between individuals and entities involved in financial transactions. By analyzing these links, the system can uncover hidden patterns and identify suspicious activities that may indicate money laundering. The proposed system leverages centrality algorithms like Degree Centrality, Closeness Centrality, Betweenness Centrality, and PageRank as key components. These algorithms assess the influence and interconnections of nodes within a network, providing a comprehensive understanding of the network's structure. By applying these centrality algorithms, the system can identify nodes that play a crucial role in potential money laundering schemes, enabling AML professionals to prioritize their investigations and allocate resources more effectively. Additionally, the system can detect anomalies and deviations from normal behavior by analyzing the interconnected nodes. In summary, this research paper addresses the limitations of conventional AML systems and proposes an advanced AML system that incorporates link analysis using deep learning techniques. By utilizing centrality algorithms, the system enhances its capability to detect suspicious activities and uncover intricate money laundering schemes. This research aims to empower financial institutions and regulatory bodies with a more effective tool to combat money laundering, ensuring the integrity and stability of the global financial industry

## Motivation

This groundbreaking research paper aims to revolutionize the field of Anti-Money Laundering (AML) by introducing an innovative approach of link analysis using deep learning techniques. By addressing the limitations of conventional AML systems, our proposed solution strives to eliminate false positives and uncover complex money laundering schemes. Through the integration of cutting-edge deep learning algorithms and our advanced AML system showcases unparalleled sophistication. We employ centrality algorithms, including Degree Centrality, Closeness Centrality, Betweenness Centrality, and PageRank, as the foundation of our system. These algorithms allow for a comprehensive examination of networks, enabling the identification of suspicious activities and connections among individuals and entities. With this research, we aim to empower financial institutions and regulatory bodies with greater capabilities to combat money laundering. By leveraging the power of link analysis, our system provides a powerful tool to detect and prevent illicit financial activities. AML professionals can benefit from the enhanced accuracy and efficiency our system offers, strengthening the global fight against money laundering. Through this groundbreaking research, we envision a future where financial systems are better equipped to identify and disrupt illicit financial activities. my motivation lies in creating a more secure and transparent global financial industry, safeguarding economies, and protecting the integrity of financial systems.

**Significance of Anti-Money Laundering**

The significance of anti-money laundering (AML) endeavors cannot be overstated within the framework of the global financial arena. These endeavors encompass a comprehensive array of legislation, regulations, and protocols meticulously devised to thwart the clandestine transformation of unlawfully acquired funds into lawful proceeds. The implementation of AML measures assumes paramount importance in up holding the credibility of financial markets and safeguarding financial institutions against becoming unwitting accomplices to criminal enterprises (C. Alexandre and J. Balsa , 2005). Not only do these efforts prove pivotal in combatting financial crimes, but they also serve as a substantial deterrent against interconnected illegal activities, including the financing of terrorism and the illicit drug trade (L. Butgereit , 2021).

**Scope**

The scope of this research paper encompasses the exploration and proposal of an advanced Anti-Money Laundering (AML) system that utilizes link analysis using deep learning techniques The primary focus is on addressing the limitations of conventional AML systems, such as high rates of false positives and the inability to uncover complex money laundering schemes. The paper delves into the potential of link analysis, which involves examining the relationships and connections between individuals and entities involved in financial transactions. By analyzing these links, the proposed system aims to uncover hidden patterns and identify suspicious activities that may indicate money laundering. Furthermore, the research paper emphasizes the use of centrality algorithms, including Degree Centrality, Closeness Centrality, Betweenness Centrality, and PageRank, within the AML system. These algorithms enhance the system's ability to identify suspicious activities by assessing the influence and interconnections within networks. Although the paper primarily focuses on the technical aspects of the proposed AML system, it also considers the broader context of the global financial industry and the role of AML efforts within it. The ultimate goal is to empower financial institutions and regulatory bodies with a more sophisticated and effective tool to combat money laundering. While the paper provides a comprehensive exploration of the proposed AML system using link analysis techniques, its scope may encompass model implementation considerations and testing, Full implementation may be further explored in future research or practical applications of the proposed system.

**Overview of the existing system**

This chapter will provide information about what money laundering is and how we can synthetically generate transaction patterns that resemble realistic fraudulent money flows. It will also include existing AML systems and current technologies and their limitations.

Money laundering "is the criminal practice of filtering ill-gotten gains, or 'dirty' money, through a series of transactions; in this way, the funds are 'cleaned'so that they appear to be proceeds from legal activities."The process of taking the proceeds of criminal activity and making them appear legal. In General, it is a process by which criminals disguise the original ownership and control of the proceeds of criminal conduct by making such proceeds appear to have derived from a legitimate source. Most anti-money laundering laws openly conflate money laundering (which is concerned with the source of funds) with terrorism financing (which is concerned with the destination of funds) when regulating the financial system. Money obtained from certain crimes, such as extortion, insider trading, drug trafficking, illegal gambling, and tax evasion is "dirty". It needs to be cleaned to appear to have been derived from non-criminal activities so that banks and other financial institutions will deal with it without suspicion. Money can be laundered by many methods, which vary in complexity and sophistication. Money launderers are big-time criminals who operate through international networks without disclosing their identity. The money laundered every year could be in the range of 600 billion Dollars to 2 trillion Dollars. This gives money launderers enormous financial power to engage coerce or bribe people to work for them. Generally, money launderers use professionals to create legal structures/ entities that act as 'front' and use them for laundering funds. Money Laundering Driving Force. Money laundering can present itself in many ways but according to (L. Butgereit , 2021). It is usually performed in three steps.

1. **Placement** Refers to the initial point of entry for funds derived from criminal activities.
2. **Layering** Refers to the creation of complex networks of transactions that attempt to obscure the link between the initial entry point, and the end of the laundering cycle.
3. **Integration** This is the movement of previously laundered money into the economy mainly through the banking system and thus such monies appear to be normal business earnings

The existing system used in Anti-Money Laundering (AML) involves several key processes to detect and prevent money laundering activities. While the specific steps may vary across institutions, the following outlines a typical process of the existing system.

- **Customer Due Diligence (CDD):** Financial institutions per- form CDD to verify the identity and assess the risk associated with their customers. This process includes collecting and verifying customer information, such as identification documents, proof of address, and beneficial ownership details.

- **Transaction Monitoring**: Financial transactions conducted by customers are continuously monitored for suspicious activities. Transaction monitoring involves the analysis of transactional data, such as transaction amounts, frequency, patterns, and parties involved, to flag any potentially illicit activities.

- **Alert Generation:** Based on predefined rules and thresholds, the system generates alerts for transactions that meet certain criteria indicative of possible money laundering or suspicious behavior. Such criteria may include large cash transactions, unusual transfer patterns, high-risk jurisdictions, or transactions involving politically exposed persons (PEPs).

- **Alert Investigation:** AML professionals review and investigate the generated alerts to determine their legitimacy. This involves analyzing additional information related to the flagged transactions, such as customer profiles, account history, and any available external data sources.

- **Suspicious Activity Report (SAR) Filing:** If an alert is deemed suspicious after investigation, a Suspicious Activity Report (SAR) is generated and filed to the appropriate regulatory authority. SARs provide detailed information about the suspicious activity, enabling further investigation and potential legal action if necessary.

- **Compliance Reporting:** Financial institutions are required to regularly report their AML activities and efforts to regulatory bodies. These reports typically include metrics, statistics, and data on the number of alerts generated, investigations conducted, and SARs filed.

- **Regulatory Compliance and Audits:** Financial institutions undergo regular audits and assessments by regulatory bodies to ensure compliance with AML regulations and guidelines. This process evaluates the effectiveness of the existing system, adherence to regulatory requirements, and the implementation of necessary controls.

It is important to note that the existing system's effectiveness varies among institutions, and some may employ additional processes or technologies, such as machine learning algorithms, anomaly detection, or data analytics, to enhance their AML capabilities. The limitations of the existing system, primarily related to high false positives and difficulty in detecting complex money laundering schemes, necessitate advancements and improvements in AML technologies and methodologies. There are three steps in this process. The first step is data collection where reports are collected from reporting entities; the second step is the analysis process where FIC analysts perform different analyses on the data collected, and finally, the dissemination.

data analytics, to enhance their AML capabilities. The limitations of the existing system, primarily related to high false positives and difficulty in detecting complex money laundering schemes, necessitate advancements and improvements in AML technologies and methodologies. There are three steps in this process. The first step is data collection where reports are collected from reporting entities; the second step is the analysis process where FIC analysts perform different analyses on the data collected, and finally, the dissemination process where the analysis result is sent to concerned organizations. This process is shown in the below diagram.

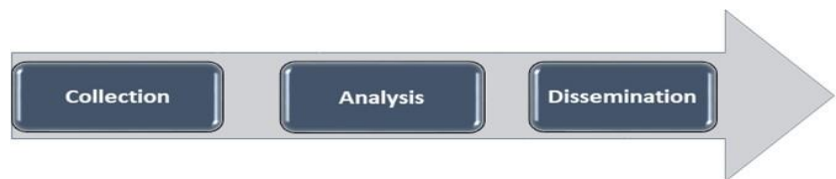

**Figure:** 1 shows Data flow in the center

**Challenges of the Existing System**

**Rule-based systems :** Financial institution needs to try and prevent and detect money laundering according to law. How this needs to be done is different in different countries. For example in the US, financial institutions needs to submit a Suspicious Activity Report (SAR) within 30 days of detecting suspicious activity. The definition of suspicious activity can be found in the Bank Security Act (BSA) where it is stated that financial institutions need to help the government detect these crimes by for example reporting transactions of over 10,000 dollars to the government (Z. Chen and W. M. Soliman , 2021) . However, in many cases, the laws and regulations do not specifically tell the financial institutions how they should do their anti-money laundering. Mostly, a rule-based system is used where transactions that meet specific criteria are flagged and thereafter manually examined.

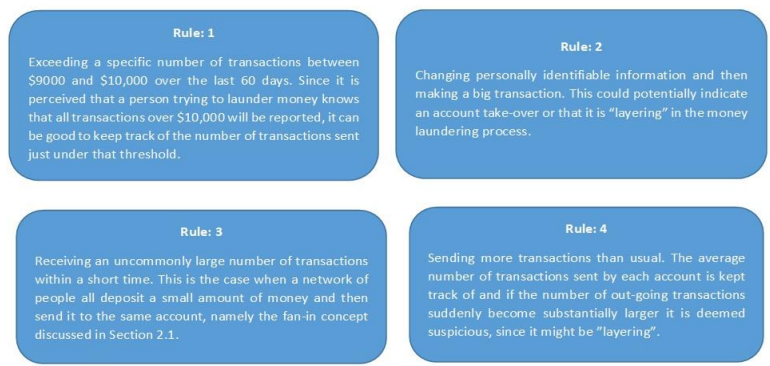

**Figure:** 2 shows Rule Based System Sample rules

These rules were taken from (K. Singh and P. Best, 2019). It should be noted that "breaking" these rules does not mean that the transaction or account is fraudulent since the rules are too broad and general. We instead have to look at cases were a single account violates several of these rules in a set time span.

**Graph-based systems**: Some of the existing AML systems are Graph-based systems and analyze graphstructured data. Hence I will explain the concept more mathematically. A graph G is a collection of nodes or vertices V, with lines known as edges E connecting them, often represented G = (V, E). The edges represent relationships between the nodes they are drawn between. In a normal graph, these relationships do not have a direction. However, the existing systems work with directed graphs, meaning that the edges have a direction with a source node and a target node. When doing deep learning on graphs it is also useful to understand the concept of adjacency. Two nodes are adjacent if they are connected with an edge. Furthermore, an adjacency matrix A is a matrix where Aij = 1 if vi and vj are adjacent. There are several ways to represent financial data as a graph but the existing system chose to have each account be represented as a node and the transactions as edges in the graph. Meaning that V = v1, . . . , vn are the n accounts and E = e1, . . . , em are them transactions. Note that there can be multiple edges (representing several transactions) between the same nodes. Moreover, the graph will be directed since each transaction has a sender and a recipient. It is also the case that each node and transaction holds a set of features to be used in the analysis

## Challenges of Existing System

In the realm of global finance, the effectiveness of anti-money laundering (AML) systems is hindered by various challenges. Traditional AML systems primarily rely on rulebased mechanisms, which often generate a significant number of false positives. Consequently, this inefficiency leads to the misallocation of resources and potential disruption of legitimate customer activities. Moreover, these systems may lack the sophistication to detect intricate money laundering schemes, which continually evolve alongside advancements in technology and global financial practices. As a result, there is a need for a nextgeneration AML framework that incorporates advanced deep-learning techniques. This new framework aims to improve the accuracy of fraud prediction, reduce false alarms, and assist domain experts in decreasing their workload (C. C. Huang and M. S. Amirrudin, 2014). The aim is to improve the accuracy of fraud prediction, reduce false alarms, and assist domain experts in decreasing their workload. This need led to the introduction of new methods of transaction monitoring using data science and machine learning techniques, with the effectiveness of these techniques being highly dependent on the quality of input features (C. C. Huang and M. S. Amirrudin, 2014). By leveraging link analysis, a method that examines associations and affiliations between entities, hidden patterns, and irregularities can be uncovered. In the context of AML, link analysis is employed to scrutinize financial transactions and identify networks of suspicious activities. Additionally, network analysis, another approach to identifying money laundering, utilizes relational data to uncover both direct and hidden connections to money laundering nodes. This involves evaluating centrality measures to determine the most important nodes in a network and applying social network metrics to assess risk profiles (Z. Gao and M. Ye , 2007). By adopting these advanced techniques, the financial industry can enhance the agility of identifying and analyzing suspicious transactions, reducing the reliance on human intervention and improving the overall effectiveness of the AML system (C. Alexandre and J. Balsa , 2005). Moreover, these systems often lack the sophistication required to detect complex money laundering schemes, which adapt to technological advancements and evolving financial practices. To address these challenges, a cutting-edge AML framework has been suggested, leveraging advanced deep learning techniques. Its objective is to enhance the accuracy of fraud prediction, minimize false alarms, and support industry experts in reducing their workload (J. Han, Y. Huang, S. Liu, and K. Towey , 2020). Evolve alongside advancements in technology and global financial practices. Consequently, there is a pressing need for more sophisticated and adaptable AML techniques that can keep up with the changing patterns and tactics employed by money launderers. The integration of innovative technologies such as link analysis, machine learning, and big data an alytics is increasingly imperative to enhance the effectiveness of AML systems.

## Dataset and Data preprocessing

This chapter includes a thorough exploration of the two data sets, to determine how to appropriately present them to the Deep learning models. First, there is a discussion about how the different data sets were synthetically generated and what makes them different from each other. Following that is a brief presentation of the AML database file configurations that the data sets came in and which features were involved. After that is a rather extensive examination of how these features can be used to engineer more useful features and which features should be dropped completely from the data sets. Lastly is a section about how we can visualize the different feature sets to get a sense for it the fraudulent and non-frudulent transactions have different feature values.

Having good quality data is important to create a satisfactory Deep learning model. However, in this thesis, we are building a model for financial transaction data, which in almost all cases are confidential. This magnifies the difficulty of getting a good data set, both concerning the quality of the data and concerning the size of the data set. The partial solution to this problem is to synthetically generate realistic data sets. By doing this data size and privacy are no longer issues but it is challenging to make the synthetic data set realistic. In 2016 the PaySim mobile money simulator was developed and presented in (J. Han, Y. Huang, S. Liu, and K. Towey , 2020). The data set consists of over 6 million transactions over a 31-day span where about 8 000 of them are fraudulent. The authors used techniques such as agent-based simulations to create a realistic synthetic dataset from a real confidential data set. However, this data set has one substantial weakness, namely that each transaction is sent from a unique account ID, meaning that we cannot infer and use information about the owner of the account. This is because each account only is involved in one transaction. Therefore we only analyze information about the transaction itself and not in the context of its environment. This is not realistic because the average account should perform more than one transaction every month. IMB saw the weaknesses and the potential of this data set and in 2018 they elaborated on it with their project AMLSim (U. G. Ketenci, T. Kurt, S. Önal, C. Erbil, S. Aktürkolu, and H. S , 2021). This data set is instead generated in two steps. First, a network of accounts, representing the nodes in the graph is created. Then the PaySim simulation, from (A. Mohan and P. Karthika , 2021) is run to generate transactions between these accounts, represented as edges. The resulting data set consists of 20,000 nodes and 866,000 edges. By doing this a more realistic proportion of accounts and transactions are generated, in conjunction with the fact that more realistic patterns between the accounts are created, these patterns are described in more detail in Section The new data simulation method also makes it possible for a single account to perform several fraudulent transactions, which can be seen in Figure 4.1 and is more realistic than fraudulent accounts only making one fraudulent transaction each. This was not possible in the old data set since each transaction had a unique sender and recipient. This data set will be used in this thesis and is from now on referred to as the AMLSim data set. Moreover, the AMLSim data set is more customizable than the PaySim data set. This is because a Github repository with the code to generate the data set is available, rather than just a CSV file. This means that the user herself can tweak and optimize different parameters in the model. This is a partial reason why IMB made the code and data set public since they welcome everyone to use the data set and help them make the data set more realistic by tweaking the many parameters in the code. One example of this is in (M. Zhang and Y. Chen , 2018) where the author evaluated 14 different configurations of the simulation. Although, this parameterization will be used in the present paper.

### Data Preprocessing Methods

Data reprocessing methods involve various techniques to modify or transform the data in order to improve its quality, remove inconsistencies, and enhance its suitability for analysis or model training. Here are some common data reprocessing methods used in the context of Anti-Money Laundering (AML):

**Data Cleaning** This process involves identifying and handling missing values, outliers, duplicates, and inconsistencies in the dataset. Missing values can be imputed or handled using appropriate techniques, outliers can be detected and processed separately, and duplicates can be removed to ensure data integrity.

**Feature Selection** Identifying and selecting relevant features from the dataset can reduce dimensionality and improve computational efficiency. Feature selection methods, such as variance thresholding, correlation analysis, or recursive feature elimination, help identify the most informative features for AML analysis.

**Data Sampling** Addressing data imbalance is crucial in AML, where suspicious transactions are usually outnumbered by legitimate ones. Techniques like oversampling (generating synthetic instances of minority class) and under sampling (removing instances from the majority class) help balance the dataset and prevent biased model training.

**Data Transformation** involves converting or normalizing the data to meet specific requirements. Common transformations include scaling numerical features to comparable scales (e.g., using min-max scaling or standardization), logarithmic transformation of skewed variables to handle skewed distributions, or encoding categorical variables into numerical representations.

**Handling Time-Series Data** AML often involves analyzing transactional or temporal data. In such cases, time-series data preprocessing techniques can be applied, such as resampling (e.g., converting data to hourly, daily, or monthly intervals), handling missing timestamps, identifying trends or seasonality, or feature engineering based on time- based variables.

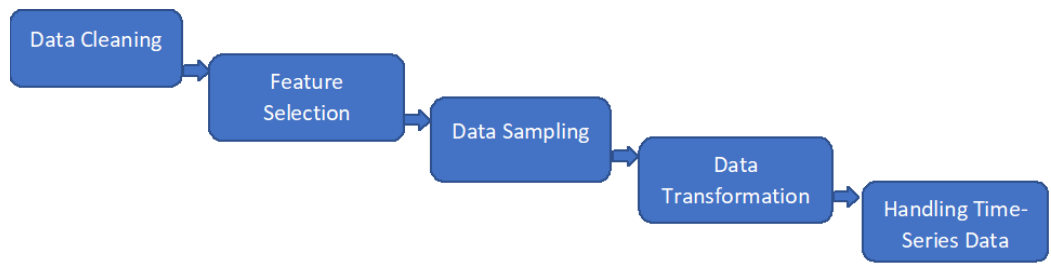

**Figure:3** shows Data Preprocessing Methods

The FICaml dataset is a compilation of customer data, transaction histories, and account details meticulously curated to fuel our advanced detection algorithms. By leveraging this comprehensive dataset, we aim to enhance our system's ability to proactively identify and address potential risks so This data set also will be used in this paper

**Methodologies**

**Proposed Algorithm and Methods**

The proposed AML system offers a comprehensive solution for combating money laundering by incorporating advanced algorithms that effectively analyze financial networks. These algorithms encompass state-of-the-art methodologies such as deep learning techniques and Centrality algortitm. In Section 4, we present a next-generation AML system that utilizes a multichannel fusion of cutting-edge deep learning techniques. We provide a detailed explanation of the system's Used Algorithm and models, highlighting how they facilitate the detection of fraudulent activities and enable effective communication of these findings to investigators. By leveraging these sophisticated techniques, the system can process large volumes of financial data with precision and accuracy. This aids in identifying potential risks and anomalies associated with money laundering activities. The integration of such state-of-the-art algorithms represents a significant step forward in combating illicit financial activities and enhancing the effectiveness of AML efforts:

**Graph Based Model**

There are several ways to represent financial data as a graph but in this paper, we chose to have each account be represented as a node and the transactions as edges in the graph. Meaning that V = v1, . . . , vn are the n accounts and E = e1, . . . , em are the m transactions. Note that there can be multiple edges (representing several transactions) between the same nodes. Moreover, the graph will be directed since each transaction has a sender and a recipient (W. L. Hamilton, R. Ying, and J. Leskovec , 2017). It is also the case that each node and transaction holds a set of features to be used in the analysis. The reason we chose graph Representation is In chapter 2 it is explained that money laundering is done in a cycle, where the money starts and ends at the same entity. This is why a graph representation of the problem is appropriate. Since representing the data as a graph makes it easier to identify relationships between accounts in a spare environment. The sparsity refers to the fact that most of the accounts only interact with a small subset of the total number of accounts. Moreover, a graph can more easily propagate the risk of a fraudulent transaction to neighboring accounts through every node's adjusting matrix. If we want to find an account's neighbors without using a graph, we would need to search through a row of a matrix of the dimension number of accounts Œ number of accounts, which is computationally fine. But if we want to find a neighbor's neighbor to a node we will have to look through this matrix multiple times making it computationally heavy. If we instead represent these relationships with edges in a graph instead of with a matrix we do not need the zeros from the matrix. we only need to specify where there is a relationship (we do not need to specify where there is not a relationship) therefore reducing the computational complexity. Henceforth, when it comes to money laundering the saying "you are your neighbors"applies, meaning that who and how an account makes transactions with other accounts is more important than information about that specific account. Furthermore, as discussed above, graphs have an easier time finding and using this information, which makes graph representations superior in this implementation (M. Zhang and Y. Chen , 2018).

**Graph Neural Networks GNNs**

Neural networks are constructed from neurons. A neuron is a function, hence it takes an input and gives an output. Mathematically we often describe neurons with

$$y = f(x) = \sigma\left(\sum_j x_i w_{ij} + b_i\right) \tag{4.1}$$

where $x_i$ is our input, $w_{ij}$ are our weights, $b_i$ is our bias and is our activation function. When these neurons are connected to a network, then we have a neural network. In Figure 3.1 we can see a representation of an arbitrary neural network, where $x_1, \ldots, x_{l'}$ are the ($j$) inputs, which in my case are the features. $h_i$ is the i:th hidden neuron in the j:th hidden layer and $y_1, \ldots, y_k$ are our outputs. The number of neurons in the hidden layers and the number of hidden layers can be varied greatly and is often an optimization task. When using a neural network we input $x$ and receive the output $y$, but what goes on in the hidden layer is unknown. This is why neural networks often are referred to as what is called black box algorithms, meaning that it is hard to see how a specific input results in a specific output. In the training stage of a neural network, we adjust the weights $w_i j$ seen in data structure to optimize the output. The optimal output is determined according to some loss function, in our case, it can be seen in Dimensionality reduction. Graph convolutional Networks are graph variants of convolutional neural networks, presented in (K. Singh and P. Best , 2019). It is a semi-supervised learning algorithm on graphs, which can be extended to supervised learning. The method was originally designed for node classification, which will be used in this thesis. However, now it can also perform link prediction (A. Mohan and P. Karthika , 2021). Before this paper, it was common to use explicit graph-based regularization on the node labels to smooth out the information over the graph, e.g. with Laplacian regularization in the loss function (T. N. Kipf and M. Welling , 2016),

$$L = L_0 + \lambda L_{\text{reg}} \tag{4.2}$$

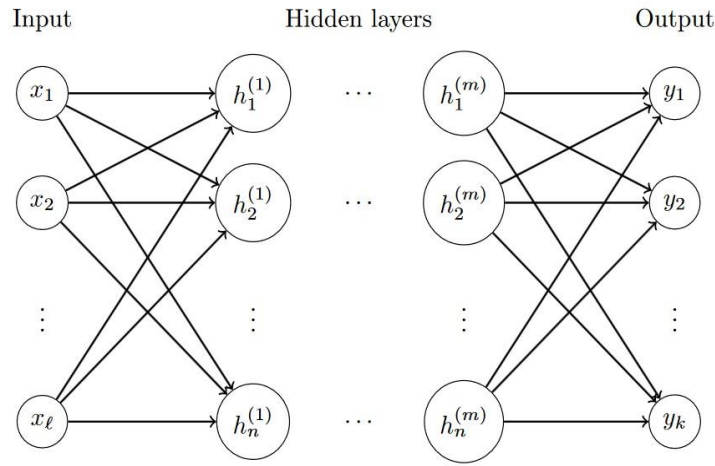

Figure 4.1 Representation of a neural network where $L_0$ is the supervised loss, is a weight factor and $L_{reg}$ is the Laplacian regularization defined as

$$L_{reg} := \sum_{i,j} A_{i,j} \|f(X_i) - f(X_j)\|^2 = f(X)^T \Delta f(X). \tag{4.3}$$

Here, $f(\cdot)$ refers to the neural network, $X$ is a matrix of the nodes, where $X_i$ are the features of node $i$, $\Delta = D - A$ is the unnormalized Laplacian of the undirected graph $G = (V,E)$ with $N$ nodes $v_i \in V$ and edges $(v_i,v_j) \in E$, $A \in \mathbb{R}^{N \times N}$ is the adjacency matrix which can be either binary or weighted, and $D$ is the degree matrix defined as $D_{ii} = \sum_j A_{ij}$.

From this representation, nodes are likely to be assigned the same label as neighboring nodes, but in some applications, edges do not mean similarity but rather some other kind of relationship. That is why Thomas N. Kipf and Max Welling extended the neural network function to also include the adjacency matrix, hence $f = f(X,A)$. By doing this, we can use gradient information about the unlabeled nodes.

**Consider the simple propagation rule:**

$$H(l+1) = \sigma(D^{-1} A H(l) W(l)), \tag{4.4}$$

where $H^{(l)} \in \mathbb{R}^{N \times D}$ is the activation matrix in layer $l$, $H^{(0)} = X$, and $W^{(l)}$ is the specific layer weights. To include the node's own features, introduce $\tilde{A} = A + I_N$, where the selfconnections are added by the identity matrix $I_N$, resulting in

$$H^{(l+1)} = \sigma(D^{-1} A \tilde{H}^{(l)} W^{(l)}). \quad (4.5)$$

Moreover, add a normalization term to the feature representation using the previously introduced degree matrix $D$, giving

$$H^{(l+1)} = \sigma(D^{-1} A \tilde{H}^{(l)} W^{(l)}). \quad (4.6)$$

Lastly, add an activation function $\sigma$:

$$H^{(l+1)} = \sigma(D^{-1} A \tilde{H}^{(l)} W^{(l)}). \quad (4.7)$$

The expression is asymmetric in terms of $i$ and $j$ since $D_{ii} = \sum_j A_{ij}$. Therefore, Kipf and Welling opted to use spectral propagation, where $D^{-1} \tilde{A}$ is replaced with $D^{-\frac{1}{2}} \tilde{A} D^{-\frac{1}{2}}$.
Resulting in the final layer-wise propagation:

$$H^{(l+1)} = \sigma\left(\tilde{D}^{-\frac{1}{2}} \tilde{A} \tilde{D}^{-\frac{1}{2}} H^{(l)} W^{(l)}\right), \quad (4.8)$$

where $\tilde{D} = \sum_j \tilde{A}_{ij}$.

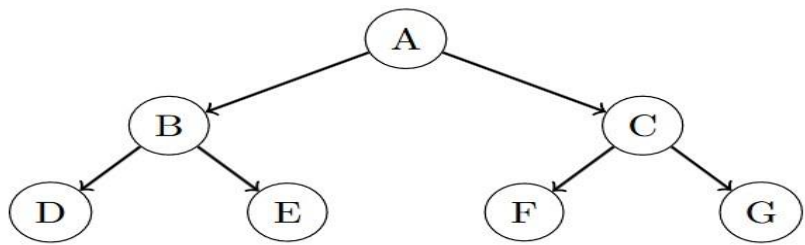

**Figure 4** Representation of a mathematical tree structure

In many cases, the GCN model only has two layers, and this is due to two reasons. Firstly, since every node is represented as a weighted average of its neighbors, adding a layer can be seen as also including the neighbors' neighbors, which means that the computational time increases exponentially with the number of layers. Secondly, there is not a positive correlation between the number of layers and the performance of the model, see, e.g.

If we let $\bar{A} = \tilde{D}^{-\frac{1}{2}}\tilde{A}\tilde{D}^{-\frac{1}{2}}$, then

$$Z = f(X,A) = \text{softmax}(\bar{A} \cdot \text{ReLU}(AXW^{(0)})W^{(1)}), \quad (4.9)$$

is a standard mathematical representation of a two-layer GCN, where $\text{softmax}(x_i) = \frac{i}{\sum_i \exp(x_i)} \exp(x)$ is the output activation.

## Data Analysis Algorithm

After implementing a Graph Convolutional Network (GCN) algorithm for financial graph link prediction, we can perform network analysis using various centrality measures and the PageRank algorithm. Below is a detailed explanation of degree centrality, betweenness centrality, closeness centrality, and the PageRank algorithm As we mentioned earlier the system has a Link analysis feature used to evaluate relationships (connections) between nodes. Relationships may be identified among various types of nodes (objects), including Accounts, Account holders, and transactions. It needs some algorithm to compute betweenness, closeness, and a degree from the link graph so we implement the PageRank algorithm.

**Degree centrality** is a useful measure for analyzing Anti-Money Laundering (AML) graphs to identify nodes that are well-connected within the financial network. In the context of AML, the nodes typically represent entities such as individuals, accounts, or organizations, and edges represent financial transactions (T. N. Kipf and M. Welling, 2016). This algorithm quantifies the number of direct connections a node possesses within the network. In the context of antimoney laundering (AML) analysis, a substantial degree of centrality may imply an entity that is extensively engaged in multiple transactions, thereby suggesting its potential role as a central hub for money laundering activities. Degree centrality is a measure of how many connections a node has. In the context of an AML graph, it can indicate how many financial transactions a node is involved in. Degree centrality is typically defined as the number of connections of a node divided by the total number of nodes in the graph (not subtracting 1). The formula for degree centrality ($C_d(v)$) is usually defined as follows:

$$C_d(v) = \frac{\text{Total number of nodes - 1}}{\text{Number of edges connected to } v} \quad (4.11)$$

Note: we Use network analysis libraries such as NetworkX in Python to calculate the degree of centrality for each node in the graph.

**Closeness centrality** is a metric employed to gauge the proximity of a node to all other nodes within a network. It serves as an indicator of the extent to which entities can readily connect with others in the network. Consequently, entities exhibiting high closeness centrality may potentially serve as facilitators within money laundering schemes. Closeness centrality measures how close a node is to all other nodes in the graph, based on the length of the shortest paths. closeness centrality can be used to identify nodes that are wellconnected to others, potentially indicating their importance or centrality within the financial network. I use network analysis libraries like NetworkX in Python to calculate closeness centrality for each node in your financial graph. the formula for Closeness centrality is usually defined as follows:

$$\text{Closeness Centrality}(v) = \frac{1}{\sum_u \text{Shortest Path Length}(v,u)} \quad (4.13)$$

**Betweenness centrality:** is a metric used to assess the frequency of a node's appearance on the shortest paths between other nodes. In the context of anti-money laundering (AML) analysis, a high betweenness centrality indicates an entity that holds significant influence in the flow of transactions. Such entities may potentially act as intermediaries in illicit activities. This centrality algorithm serves as a valuable tool in identifying key players involved in money laundering, aiding in the detection and prevention of financial crimes.Betweenness centrality identifies nodes that act as bridges along the shortest path between other nodes in the graph. The formula for Betweenness centrality usually defined as follows:

$$\text{Betweenness Centrality}(v) = \sum_{s \neq v \neq t} \frac{\sigma_{st}(v)}{\sigma_{st}} \quad (4.12)$$

where $\sigma_{st}$ is the total number of shortest paths from node s to node t, and $\sigma_{st}(v)$ is the number of those paths that pass through node v.

**PageRank:** PageRank measures the transitive (or direction) influence of nodes and is a variant of the Eigenvector Centrality algorithm. Eigenvector Centrality can be used on undirected graphs, whereas the PageRank algorithm is more suited to directed graphs [8]. PageRank is the best algorithm in an anomaly or fraud detection system in the financial sector. So we proposed this algorithm to implement for graph link analysis. The PageRank algorithm plays a pivotal role in the context of anti-money laundering (AML) systems, particularly in graph link analysis for anomaly or fraud detection in the financial sector. This algorithm is a variant of the Eigenvector Centrality algorithm, which is suitable for undirected graphs, while PageRank is more apt for directed graphs and The PageRank of a node v is calculated iteratively based on the PageRank values of its incoming links. The formula is more complex but it can be implemented using network analysis libraries.PageRank algorithm equation:

$$PR(A) = (1-d) + d\left(\frac{PR(T_1)}{C(T_1)} + \ldots + \frac{PR(T_n)}{C(T_n)}\right) \quad (4.10)$$

## Data Analysis Methods

**Link Analysis/Visual Link Methods**

- Link analysis feature used to evaluate relationships (connections) between nodes. Relationships may be identified among various types of nodes (objects), including Accounts, Account holders, and transactions (Choo et al., 2014).
- Is there redundant personal information or suspicious associations with high-risk entities?
- Peer group anomaly detection.
- Correspondent banking scenarios and a relationship grid for assessing details associated with involved parties.

    - Link(s) among different accounts

    - Link(s) between person and companies

    - Link between persons

    - Link between companies

## Data Analysis Algorithm

As we mentioned earlier the software will have a Link analysis feature used to evaluate relationships (connections) between nodes. Relationships may be identified among various types of nodes (objects), including Accounts, Account holders, and transactions. It needs some algorithm to compute beteweeness, closeness, and a degree from the link graph so we implement the PageRank algorithm (Gao & Ye, 2007)].

**PageRank algorithm that solves the following equation:**

---

**Algorithm 1** PageRank Algorithm

---

$$PR(A) = (1 - d) + d(PR(T1)/C(T1) + ... + PR(Tn)/C(Tn)) \quad (1)$$

---

- In this context, an Account A is assumed to have Transactions T1 to Tn pointing to it.
- The damping factor d is set between 0 (inclusive) and 1 (exclusive), typically at 0.85. This factor represents the probability of randomly following a link from a given node.
- C(A) is defined as the number of links going out of Account/Transaction
- A. This is used to normalize the contribution of each transaction.

# Implementation

The implementation of the next-generation AML system involves integrating it with existing financial reporting frameworks. This is done by seamlessly integrating the system with financial databases and reporting tools. The implementation process also focuses on prioritizing data security and compliance to ensure that sensitive financial information is protected. To ensure successful implementation, user training and support are provided to ensure that end-users are familiar with the system and can effectively utilize its features. The prototype application AML Link serves as a proof of concept for the implementation strategy. It has been tested and verified using sample data to demonstrate its capabilities. Although AML Link is currently a prototype and not a commercial application for auditors and other end users, its development showcases the potential of the proposed next-generation AML framework. As the system progresses, it will be refined and enhanced to meet the specific needs of auditors and other stakeholders in the financial industry (Singh & Best, 2019).

**Data Source**

We used a sample database named **AMLSim** database this database contains the following tables as:

- Account Information Table
- Account Table
- Transaction Information Table
- Beneficiary Information Table
- Reporting Entity Table
- Reporting Entity Branch Table

These tables contain Suspicious Transaction Report/STR, Cash Trans- action Repot/CTR, and Cross Border Transaction Report/CBTR

**Note**: we used SQL Database to store and organize our transactional data

**Operate Anti-Money Laundering System**

**Adding Target Entity from Entity Palette**

We can add any options from entity palette as a Target node

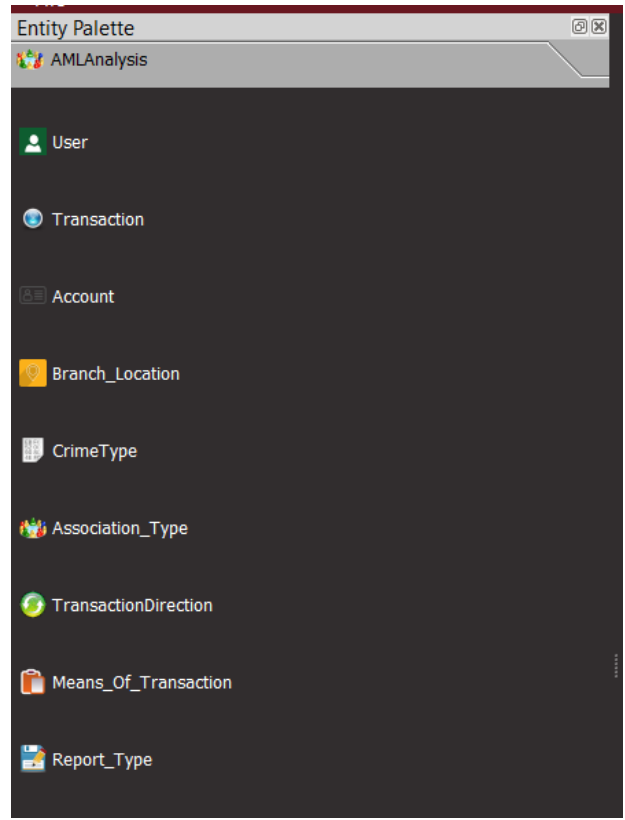

- To add Target From reporting entity palette write click on one of the target listed in the palette and click on add button then the will provide target entry box

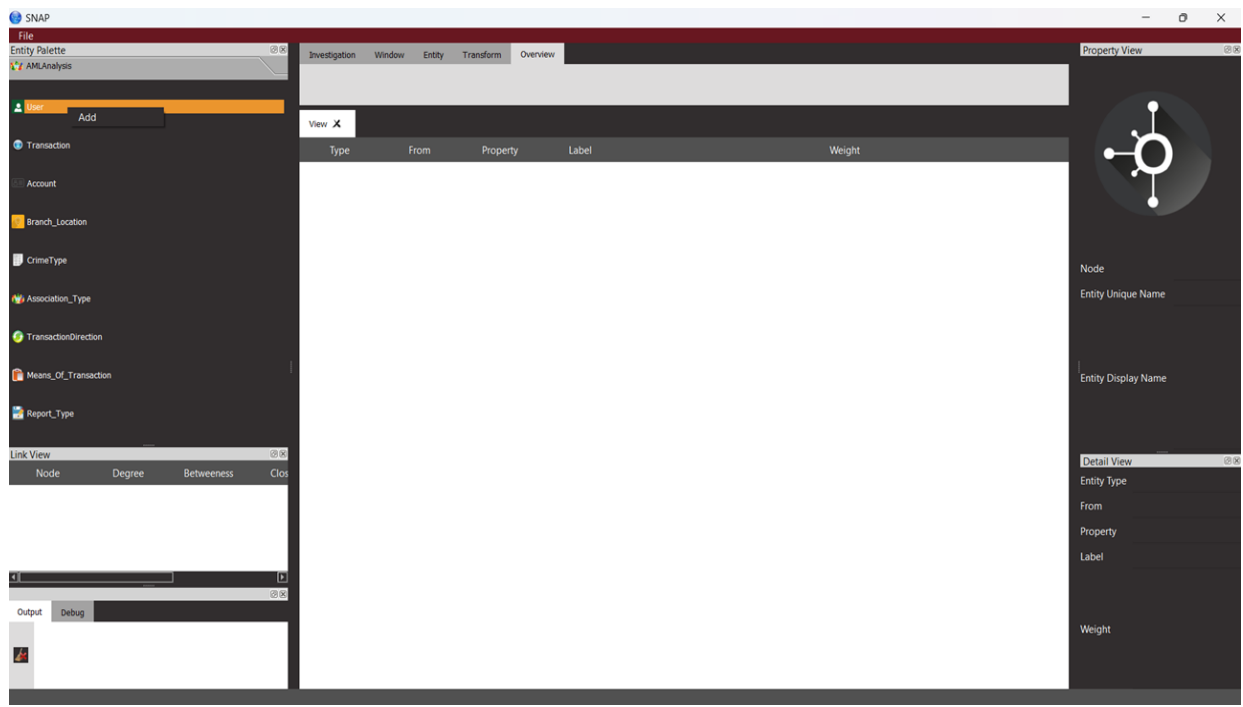

- Enter Target Full Name, the target name will be individual name or entity or company name of account owner

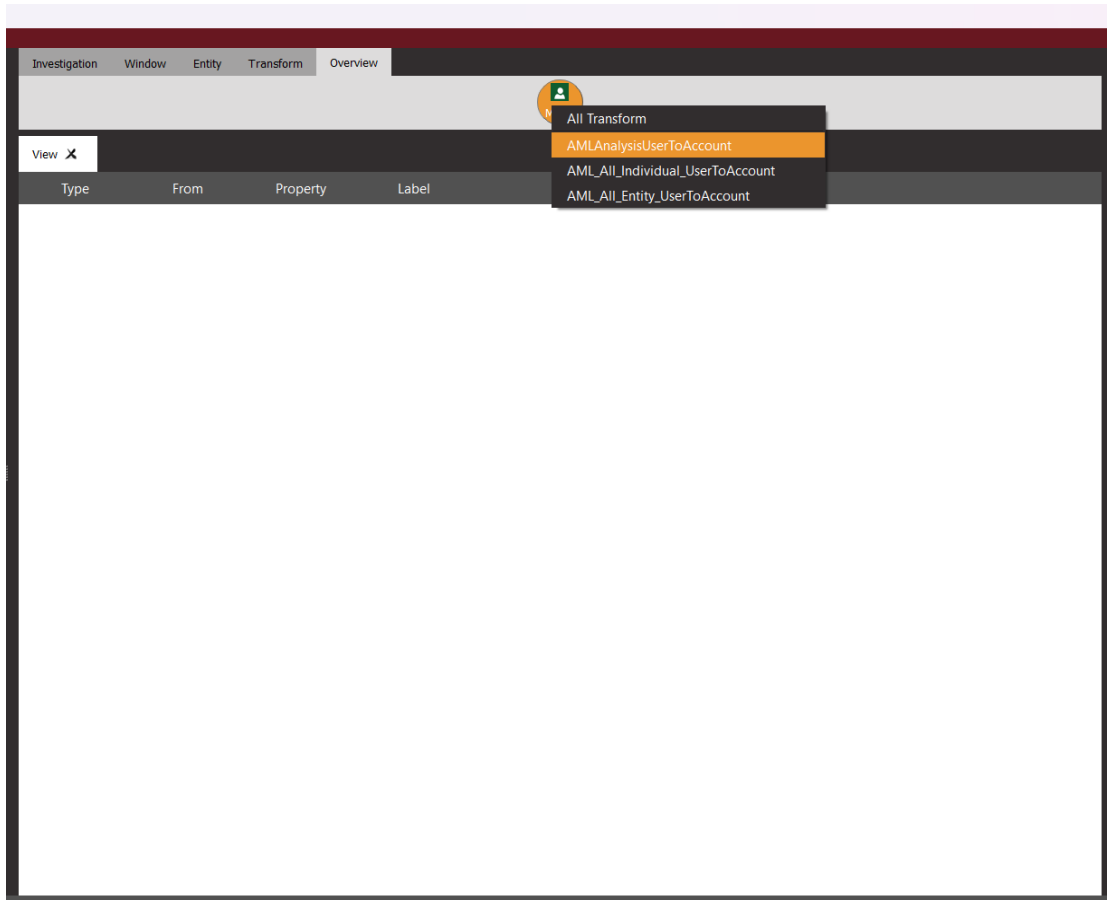

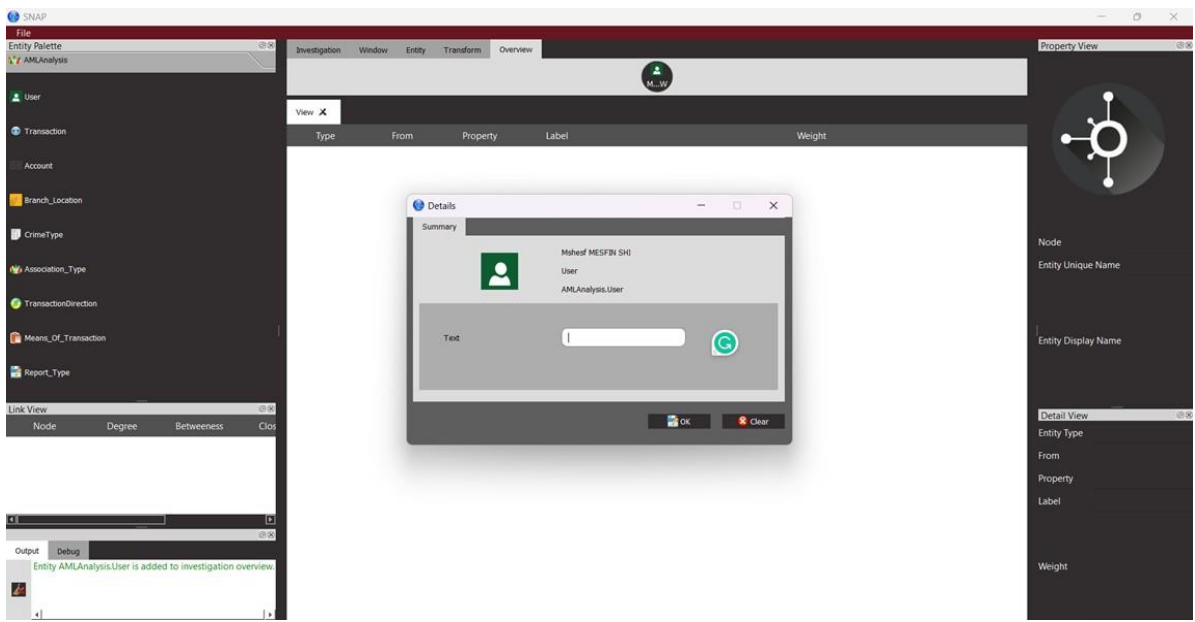

- link the target account to different scenarios by using python transform script for example
    - user — account
    - individual User — account
    - entity user — account
    - we can link this specific user to all transforms

**Link Analysis graph and textual view**

The system provides link node visualization and textual information about each node and edge we can investigate further by right-clicking on the node until the leaf node or last depth

## Conclusion

The development and implementation of an advanced Anti-Money Laundering (AML) system utilizing the Deep learning technique mark a significant evolution in combating financial crimes. This system, with its integration of The GCN algorithm almost managed to achieve a perfect classification of the test set with only two misclassifications one false negative and one false positive, resulting in an accuracy of 0.999. Because of this we do not have the same decision concerning the imbalance between false negatives and false positives. The parameters in the model will therefore not have to be tweaked for the specific company's or institutions implementation. It would have been interesting to see if this difference is due to a smaller test set or if the GCN method more accurately can find the positive cases. Conclusively, we would strongly recommend the GCN algorithm for detecting money laundering on graph structured financial transaction data because of its accuracy computation speed. In conclusion the GCN model outperforms the XGBoost algorithm, which confirms our research question. Hence it is possible to successfully combine a GCN with a NENN to make a graph learning algorithm that uses edge features in order to detect money laundering from transactional data. We also recommend Centrality algorithms like Degree Centrality, Closeness Centrality, Betweenness Centrality, and PageRank, offers a more dynamic and comprehensive approach to identifying and analyzing suspicious activities in financial transactions. The core strength of this system lies in its ability to map and interpret complex transaction networks, thus uncovering patterns indicative of money laundering that traditional systems may miss. Furthermore. Its seamless integration with existing financial databases and reporting tools, alongside rigorous adherence to data security and regulatory compliance standards, makes it a robust solution for financial institutions. The system not only enhances the detection capabilities of money laundering activities but also significantly improves operational efficiency by reducing false positives and streamlining the detection process. Looking forward, the potential integration of emerging technologies like artificial intelligence, machine learning, and block chain is set to further revolutionize AML efforts. These advancements promise enhanced accuracy, efficiency, and predictive capabilities, thereby fortifying the financial sector against increasingly sophisticated criminal activities. In summary, the proposed AML system represents a critical step forward in safeguarding the integrity and stability of the global financial landscape. Its adoption is essential for a more secure, compliant, and efficient financial environment, benefiting not just individual institutions but the economy at large. Continuous innovation and adaptation in AML technologies re- main key to staying ahead in the ever-evolving battle against financial crimes.

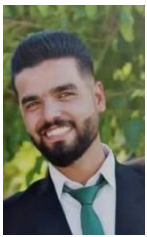

MASHKHAL ABDALWAHID SIDIQ
College of Software EngineeringNankai University
mashkhal.abdwl@mail.nankai.edu.cn.

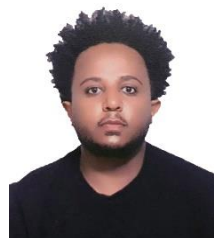

YIMAMU KIRUBEL WONDAFEREW
College of Software EngineeringNankai University
kirubel.wondaferew1978@gmail.com

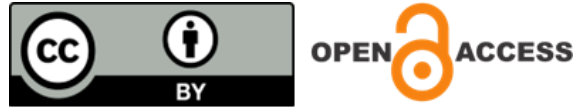

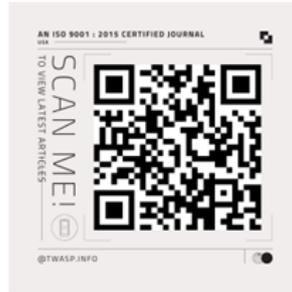